\documentclass{article}
\usepackage{VMDLinear,times}
\usepackage{graphicx}
\usepackage{natbib}
\usepackage{multirow}
\usepackage{parskip}
\graphicspath{ {images/} }
\DeclareUnicodeCharacter{2061}{}

\usepackage{amsmath,amsfonts,bm}

\def\eqref#1{equation~\ref{#1}}
\def\1{\bm{1}}

\DeclareMathAlphabet{\mathsfit}{\encodingdefault}{\sfdefault}{m}{sl}
\SetMathAlphabet{\mathsfit}{bold}{\encodingdefault}{\sfdefault}{bx}{n}

\usepackage{hyperref}
\usepackage{url}

\title{Variational Mode Decomposition and Linear Embeddings are What You Need For Time-Series Forecasting}

\author{Hafizh Raihan Kurnia Putra \\
Department of Informatics Engineering\\
Brawijaya University\\
Malang, Jawa Timur, Indonesia \\
\texttt{hachiman170@student.ub.ac.id} \\
\And
Novanto Yudistira \\
Department of Informatics Engineering\\
Brawijaya University\\
Malang, Jawa Timur, Indonesia \\
\texttt{yudistira@ub.ac.id} \\
\AND
Tirana Noor Fatyanosa \\
Department of Informatics Engineering\\
Brawijaya University\\
Malang, Jawa Timur, Indonesia \\
\texttt{fatyanosa@ub.ac.id} \\
}

\iclrfinalcopy % Uncomment for camera-ready version, but NOT for submission.
\begin{document}
\maketitle
\begin{abstract}
Time-series forecasting often faces challenges due to data volatility, which can lead to inaccurate predictions. Variational Mode Decomposition (VMD) has emerged as a promising technique to mitigate volatility by decomposing data into distinct modes, thereby enhancing forecast accuracy. In this study, we integrate VMD with linear models to develop a robust forecasting framework. Our approach is evaluated on 13 diverse datasets, including ETTm2, WindTurbine, M4, and 10 air quality datasets from various Southeast Asian cities. The effectiveness of the VMD strategy is assessed by comparing Root Mean Squared Error (RMSE) values from models utilizing VMD against those without it. Additionally, we benchmark linear-based models against well-known neural network architectures such as LSTM, BLSTM, and RNN. The results demonstrate a significant reduction in RMSE across nearly all models following VMD application. Notably, the Linear + VMD model achieved the lowest average RMSE in univariate forecasting at 0.619. In multivariate forecasting, the DLinear + VMD model consistently outperformed others, attaining the lowest RMSE across all datasets with an average of 0.019. These findings underscore the effectiveness of combining VMD with linear models for superior time-series forecasting. Code is available at this repository: \url{https://github.com/Espalemit/VMD-With-LTSF-Linear.git}
\end{abstract}

\section{Introduction}
\label{sec:introduction}
Time-series information forecast plays a significant part in estimating future occasions, particularly with the integration of machine learning methods. Machine learning models can be prepared to realize long expectation precision and low error rates by leveraging information from datasets. 

Achieving optimal results with machine learning models requires a several preprocessing steps such as data cleaning, data decomposition, hyperparameter tuning, and etc. Additionally, selecting the appropriate time-series forecasting method from a the available strategies is crucial.

One widely used algorithm in time series forecasting is the Transformer, that is well known for its performance in Natural Language Processing (NLP), speech recognition, and computer vision assignments. However, \citet{Zeng2022} has raised questions about the adequacy of Transformers in time-series forecast, inciting comparative execution assessments against easier linear-based calculations such as Linear, NLinear, and DLinear. Comparative tests have appeared that direct models frequently outperformed Transformer-based models like FEDFormer, AutoFormer, Informer, LogTrans, and PyraFormer in terms of Mean Squared Error (MSE) and Mean Absolute Error (MAE) \citep{Zeng2022}.

Furthermore, data decomposition strategies essentially affect forecast results. Variational Mode Decomposition (VMD) is one such strategy competent of lessening data instability \citep{Dragomiretskiy2014}. The performance of VMD in time-series information forecast was highlighted in a 2021 study about comparing Recurrent Neural Network (RNN)-based models under different scenarios with PM2.5 air quality data over a few cities in China: 1) without decomposition method, 2) using Empirical Mode Decomposition (EMD), and 3) with VMD utilization. Based on the results, it was shown that VMD consistently achieved the least Root Mean Squared Error (RMSE), especially with Bidirectional LSTM model \citep{Zhang2021}.

Given VMD's capability to diminish data volatility and good performance of linear-based models, there exists potential for more compelling time-series data forecasting indeed with less stable data. These models will be compared against other deep learning models such as RNN, Long Short Term Memory (LSTM), and Bidirectional LSTM (BiLSTM), which are less difficult however broadly utilized in time-series forecasting researches.

motivated by the reasons outlined, this aims to investigate "Variational Mode Deterioration with Straight Embeddings for Successful Time-Series Information Expectation." The goal of employing VMD for data decomposition and evaluating simplistic linear models (Linear, NLinear, and DLinear) alongside RNN-based models is to develop a predictive framework capable of achieving minimal error rates even under volatile data conditions.

\section{Related Works}
\label{sec:related}
\subsection{Time-Series Forecasting Models}
\textbf{Linear models}. Linear model was originally used as a regression tool in statistics that predict outcomes based on linear relationship between independent variables and dependent variables. Its application on machine learning and pattern recognition were discussed by \citet{Bishop2006} in a broader context. Their simplicity and interpretability made them valuable for many tasks, though they can struggle with capturing complex, non-linear relationships. In a research conducted by \citet{Zeng2022}, linear models were improved into NLinear and DLinear that are capable to compete with many transformer based models in time-series forecasting. Another linear-based model for long term time series forecasting is RLinear which was introduced back in 2023 \citep{Li2023RevisitingLT}. RLinear utilized Reversible Instance Normalization (RevIN) and Channel Independent (CI) strategy to improve overall forecasting performance. RevIN operates by normalizing each instance of the data independently which can lead to improved convergence rate and reduced overfitting \citep{kim2021reversible}. Meanwhile CI is a strategy used in multivariate time-series forecasting that normalizes the data separately for each feature \citep{han2024capacity}.

\textbf{Neural networks}. Neural network is a more robust algorithm in machine learning that was inspired by the human nervous system, consist of layers of interconnected nodes (neurons) and are designed to model complex, non-linear relationships. These models are trained using backpropagation to adjust weights and improve prediction results \citep{Aggarwal2018}. They can range from simple feedforward networks to deep architectures with many layers such as Recurrent Neural Network (RNN). However due to its limitations on long term forecasting, Long Short Term Memory (LSTM) was introduced as an improvement from RNN. LSTM still used RNN architecture but it was designed to handle long-term dependencies in data \citep{Sherstinsky_2020}. 

\textbf{Transformer} As a more recent advancement, an algorithm with self-attention mechanism was introduced, self-attention allows the model to weigh the importance of different parts of an input sequence relative to each other. This architecture are called Transformer and it's capable of processing all elements simultaneously, rather than sequentially, enhancing both efficiency and performance in capturing complex dependencies \citep{vaswani2023}. Transformer is still developed to this day with application on other fields such as AutoFormer and Reformer for time-series forecasting. AutoFormer operates by decomposing the data into trend and seasonal then applied together with Auto-Correlation \citep{wu2021autoformer}. Meanwhile, Reformer utilizes local attention on a fixed window together with reversible layers to reduce the complexity of computation \citep{kitaev2020reformer}. Transformer can be further advanced by unifying the statistics of each input then converting the output with restored statistics, and also by recovering intrinsic non-stationary information into temporal dependencies, those advancement were done in Stationary Transformer \citep{Koyejo2022}. Another advancement that transforms the limitations of 1D temporal variation into the 2D space to allow temporal evolution of one feature to be affected by the other \citep{wu2023timesnet}. In 2023, PatchTST were introduced, a Transformer that utilizes Channel Independance (CI) and segmentation of time series inputs into several subseries-level patches. Each patch represents a portion of the original data \citep{Yuqietal-2023-PatchTST}.

\textbf{Large Language Models}. Large Language Models are models that were designed to understand, work, and generate with various human language. However, it was also recently developed to work with images for Computer Vision (CV) by utilizing its robustness in pattern recognition and reasoning \citep{mirchandani2023}. However, well-known LLM such as GPT and BERT can be used in forecasting time series data by fine-tuning it based on temporal data and also unifying their framework so that it can be used in diverse sequential task (classification, anomaly detection, forecasting, etc.) \citep{zhou2023onefitsall}. Time-LLM were developed in this year, a time series LLM model were developed to work with time-series data by reprogramming time series inputs into text prototype representations and guiding LLM's reasoning by augmenting the input context with declarative prompts \citep{jin2023time}. It was proven competitive against many recently advanced time series forecasting method such as GPT4TS, DLinear, PatchTST, etc. 

\subsection{Data Mode Decomposition}
Data mode decomposition is a technique used to decompose complex time series data into many data modes or Intrinsic Mode Function (IMF) with less complexity. Empirical Mode Decomposition (EMD) is one of data decomposition method that decomposes a signal into a set of intrinsic mode functions (IMFs) and a residual trend, capturing components of varying frequency ranges adaptively. EMD operates by iteratively sifting the data to identify and extract these IMFs, making it particularly effective for non-stationary and nonlinear time series data \citep{Huang1998}. However, EMD can be sensitive to noise and may suffer from mode mixing, where different modes are incorrectly combined. 

To address those limitations, Variational Mode Decomposition (VMD) was introduced as a more robust data decomposition method. VMD decomposes a signal into modes by solving a variational problem that seeks to minimize the bandwidth of each mode while ensuring they are as distinct as possible from each other. It utilized Lagrange function and Alternate Direction Method of Multipliers (ADMM) \citep{Dragomiretskiy2014}. Lagrange function uses \(\lambda\) penalty term to enforce constrain for decomposed components. Lagrange is a quite complex function that can be simplified by implementing ADMM to split the function into several vital parts of function. ADMM works by iterating those functions until it reaches convergence \citep{Boyd2011}. This method is less prone to noise and mode mixing issues compared to EMD, as it uses a principled optimization approach to separate modes in the frequency domain. Dynamic Mode Decomposition (DMD) is an another data decomposition method that is capable of extracting important dynamic characteristics of the data. DMD decomposes the data into several data modes or IMFs by utilizing matrix analysis and singular value decomposition \citep{Kutz2016}.

\section{Model}
\label{sec:model}

\subsection{VMD-Linear Framework}

\begin{figure}[htbp]
    \begin{center}
    %\framebox[4.0in]{$\;$}
    \includegraphics[width=0.5\linewidth]{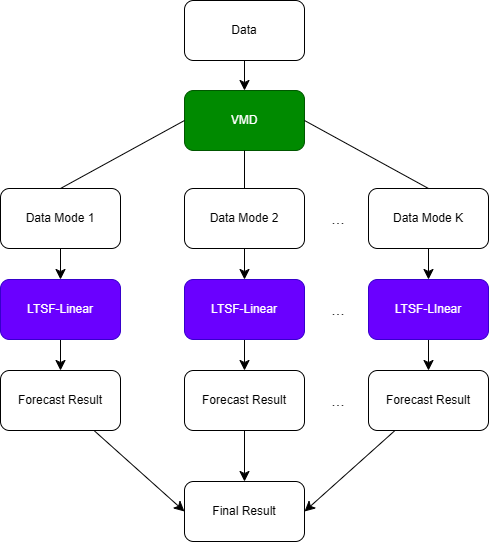}
    %\fbox{\rule[-.5cm]{0cm}{4cm} \rule[-.5cm]{4cm}{0cm}}
    \end{center}
\caption{VMD-Linear Framework}
\label{figure:framework}
\end{figure}

The structure of VMD-Linear shown in Figure \ref{figure:framework} was inspired by VMD-BLSTM framework introduced by \citet{Zhang2021}. The goal is to reduce data bandwidth in all IMF or data modes with the sum of all IMF equals to that of the original data. LTSF-Linear models will be assigned on each IMF for time-series forecasting. As the final step, the individual forecasts will be aggregated to produce the overall forecast result.

\subsection{Variational Mode Decomposition}

Variational Mode Decomposition (VMD) is a method commonly used to decompose complex signals into several k parts of signals (modes) that have been modulated in terms of amplitude and frequency, also known as Intrinsic Mode Function (IMF).

\begin{equation}
\label{1eq}
    IMF\; or\; u_k(t) = A_k(t) \cos⁡ (\phi _k (t))
\end{equation}

The symbol \(\phi_k(t)\) is a non-decreasing function (can be considered a wave) with the condition that \(\phi_k(t) \geq 0\), and \(A_k(t)\) is the envelope of the wave \(\phi_k(t)\). All IMF formed must satisfy Equation 1. VMD applies Hilbert transform to obtain the analytic frequency spectrum (\(Au_k\)) of each mode \(u_k\). In the transformation process, each mode is multiplied by the center frequency (\(e^{-jw_k t}\)) to shift its frequency spectrum towards the baseband \citep{Dragomiretskiy2014}.

\begin{equation}
\label{2eq}
    Au_k(t) = [\delta(t)+ \frac{j}{\pi t}] * u_k (t) e^{-jw_k t}
\end{equation}

The Dirac distribution is symbolized as \(\delta\) and j is the imaginary number satisfying \(j^2 = -1\). The symbol \(w_k\) in the center frequency (\(e^{-jw_k t}\)) denotes the central pulse/vibration of each mode. The signal's bandwidth will be estimated using \(H^1\) Gaussian Smoothness of the demodulated signal. In this case, the estimation is performed by squaring the \(L^2\) normalization of the gradient.

\begin{equation}\label{3eq}
    min_{u_k ,w_k}⁡\{\Sigma _k \parallel \partial _t  [(\delta(t)+ \frac{j}{\pi t})  \ast u_k (t)] e^{-jw_k t} \parallel _2^2\} 
\end{equation}

Equation 3 only applies when \(\Sigma_k\; u_k  = original\;signal\).The main objective of VMD is to minimize the bandwidth around the center frequency in each mode \(m_k\) while maintaining the condition that the sum of all modes equals the original signal. It is recommended to use Lagrange multiplier (\(\lambda\)) and quadratic penalty. The quadratic penalty is used to address additional Gaussian noise. In noise-free conditions, infinite weights are required to enforce strict data accuracy. The Lagrange multiplier is a way to impose strict constraints. By combining the strict constraint imposition of the Lagrange multiplier and the quadratic penalty (\(\alpha\)) with finite weights, the enhanced Lagrangian function L is formulated in Equation 4 to transform the constrained problem.

\begin{equation}\label{4eq}
\begin{split}
    & L ({u_k },{w_k },\lambda) = \alpha \Sigma_k\parallel \partial _t  [(\delta(t)+ \frac{j}{\pi t})  \ast u_k (t)] e^{-jw_k t} \parallel _2^2 + \\
    & \parallel f(t)- \Sigma_k u_k (t)\parallel _2^2 + [\lambda(t),f(t)- \Sigma_k u_k (t)] \\
\end{split}
\end{equation}

The minimization problem in Equation 4 can be addressed using the Lagrange function L by employing the Alternate Direction Method of Multipliers (ADMM) iteratively on \(u_k\), \(w_k\), and \(\lambda\). Iteration will stop if the condition is met: \(\Sigma _k (\parallel m_k^{n+1} - m_k^n \parallel _2^2 /  \parallel m_k^n\parallel _2^2) < \varepsilon\) where \(\varepsilon\) is the threshold value.

\begin{equation}\label{5eq}
    \widehat{u} _k^{n+1}  = \frac{\hat{f} (w)- \Sigma_{i \neq k}\; \widehat{u}_i (w)+ \frac{\widehat{\lambda} (w)}{2}}{1+2\alpha (w - w_k)^2 }
\end{equation}

\begin{equation}\label{6eq}
    \widehat{w} _k^{n+1}  = \frac{\int_0^\infty w |\widehat{u} _k (w)|^2  dw}{\int_0^\infty |\widehat{u} _k (w)|^2  dw}	
\end{equation}

\begin{equation}\label{7eq}
    \hat{\lambda} ^{n+1} (w) = \hat{\lambda} ^n  (w) + \tau(\hat{f}(w) - \Sigma_k\widehat{u} _k^{n+1} )
\end{equation}

For multivariate forecasting, the main idea still identical but most variables is represented in vectors. Thus, increasing the complexity and computational demands \citep{Rehman2019}.

\subsection{Long-term Time Series Forecasting Linear}

Long-term Time Series Forecasting Linear or LTSF-Linear is a sequence of simple one-layer linear models consisting of Linear, NLinear, and DLinear \citep{Zeng2022}. LTSF-Linear models already support multivariate forecasting by alternating the number of channels in the parameter. LTSF-Linear models is more efficient in terms of time and resource due to having no more than two linear layers, allowing training with smaller memory and fewer parameters possible. Linear model is the simplest model with a simple one layer linear model, unlike NLinear and DLinear.

\begin{figure}[h]
    \begin{center}
    %\framebox[4.0in]{$\;$}
    \includegraphics[width=0.5\linewidth]{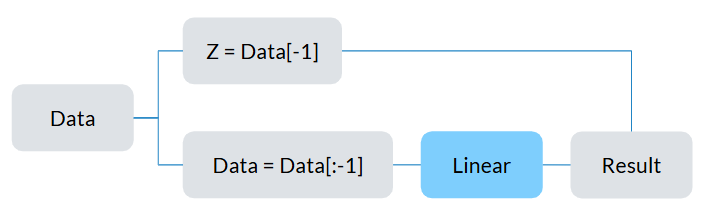}
    %\fbox{\rule[-.5cm]{0cm}{4cm} \rule[-.5cm]{4cm}{0cm}}
    \end{center}
\caption{NLinear Framework}
\label{figure:nlinear}
\end{figure}

\textbf{NLinear} As shown in Figure \ref{figure:nlinear}, NLinear model is used to anticipate distribution shifts in the data by reducing the last input passed to the Linear layer and then adding back the reduced input before making the final prediction.

\begin{figure}[h]
    \begin{center}
    %\framebox[4.0in]{$\;$}
    \includegraphics[width=0.5\linewidth]{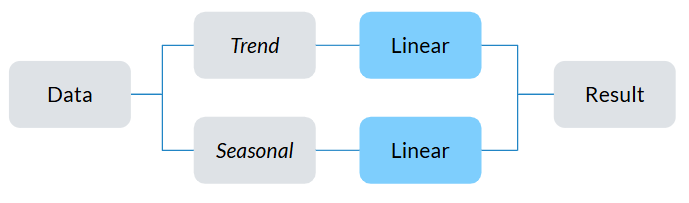}
    %\fbox{\rule[-.5cm]{0cm}{4cm} \rule[-.5cm]{4cm}{0cm}}
    \end{center}
\caption{DLinear Framework}
\label{figure:dlinear}
\end{figure}

\textbf{DLinear} DLinear enhances the simple linear model by utilizing moving average to split the data into Trend and Seasonal with each going through their seperate linear layer. This model should be used if there is a clear trend in the data. DLinear framework is shown in Figure \ref{figure:dlinear}.

\section{Experiments}
\label{sec:experiment}

In this section, we present experiments using LTSF-Linear models and few neural network models such as Recurrent Neural Network (RNN), Long Short Term Memory (LSTM), and Bidirectional LSTM (BLSTM) on 13 real-world datasets. In this research, we compare each model performance using RMSE values.

\subsection{Dataset}

We used 13 different real-word dataset obtained from various sources in the internet. The datasets are described as below:
\begin{enumerate}
    \item \textbf{ETTm2 Dataset}, this dataset were compiled by Beijing Guowang Fuda Science and Technology Department \citep{haoyietal-informer-2021}. It consist electricity distribution dynamics and were captured in 15-minutes intervals with a total of 69,680 data points.
    \item \textbf{Wind Turbine}, The following data contains information about a certain wind turbine in Turkey that were captured using the SCADA system \citep{Erisen2019}. Wind Turbine related data such as wind speed, wind direction, and power generated were captured in 10-minute intervals with a total of 50,500 data points.
    \item \textbf{M4 Dataset}, The M4 dataset were originally used in M4 Competition as a benchmark to evaluate various models in real word scenarios. It consist many features and can be used for multivariate forecasting \citep{MAKRIDAKIS202054}. This dataset were recorded in 7-days intervals.
    \item \textbf{SEA Air Quality Dataset}  We also used various real-world Air Quality Index (AQI) datasets from many cities recorded from 27 January 2022 to 22 Mei 2023 \citep{AQI2023}. Those cities were Central Singapore, Banjarbaru, Batam, Jakarta, South Jakarta, Jambi, Malang, Medan, Samarinda, and Semarang. 
\end{enumerate}
ETTm2, Wind Turbine, and M4 dataset were used because of its relatively high variance in their features \citep{Arthur2024}. Those 3 datasets will be used in multivariate forecasting together with Central Singapore AQI dataset. As for univariate forecasting, all dataset will be used including mentioned multivariate datasets.

We divide the dataset into training data and test data with a certain ratio. Dataset fewer than 500 rows will be split with 90:10 ratio, and dataset with more than 10000 rows will be trimmed to the first 10,000 rows then split with 80:20 ratio. We conducted two different testing scenario based on number of features, namely univariate and multivariate. On univariate testing scenario, we only used one feature on each dataset in this testing scenario. Meanwhile, multivariate testing scenario used 4 datasets that contain multiple feature such as ETTm2, Wind Turbine SCADA, M4 Competition, and Central Singapore Air Quality dataset.

\subsection{Experimental Protocols}

First, dataset were cleaned then normalized by using Standard Scaler. After normalization, variational mode decomposition were implemented on the dataset and will create K data modes or IMFs. Each data modes will be going through separate forecasting model, then the forecasting results of each modes is summed together to attain the final forecasting result. We used LASSO regularization in loss calculation to avoid vanishing gradients that could happen while training the models \citep{Mairal2012}. RMSE were used as a metric evaluation method for emphasizing each model's impact on reducing error rates \citep{Hodson2022}.

\begin{table}[!h]
\caption{Multivariate RMSE Results}
\centering
{\scriptsize
\setlength\tabcolsep{1.5pt}
\begin{tabular}{c|cl|cl|cl|cl|cl|cl}
\multirow{2}{*}{\textbf{Dataset}} & \multicolumn{2}{c|}{\textbf{Linear}}  & \multicolumn{2}{c|}{\textbf{DLinear}}          & \multicolumn{2}{c|}{\textbf{NLinear}} & \multicolumn{2}{c|}{\textbf{LSTM}} & \multicolumn{2}{c|}{\textbf{BiLSTM}} & \multicolumn{2}{c}{\textbf{RNN}}  \\ \cline{2-13} & \multicolumn{1}{l}{No VMD} & VMD   & \multicolumn{1}{l}{No VMD} & VMD            & \multicolumn{1}{l}{No VMD} & VMD   & \multicolumn{1}{l}{No VMD} & VMD   & \multicolumn{1}{l}{No VMD}  & VMD    & \multicolumn{1}{l}{No VMD} & VMD   \\ \hline
\begin{tabular}[c]{@{}c@{}}{\tiny M4 Weekly}\end{tabular} & 1,041                      & 0.102 & 1,063                      & \textbf{0.040} & 1,116                      & 0.064 & 1,070                      & 0.092 & 1,126                       & 0.089  & 1,092                      & 0.080 \\
{\tiny Central Singapore}                                              & 1,203                      & 0.098 & 1,221                      & \textbf{0.024} & 1,331                      & 0.051 & 1,165                      & 0.124 & 1,218                       & 0.203  & 1,204                      & 0.077 \\
{\tiny Wind Turbine}                                                      & 1,397                      & 0.007 & 1,404                      & \textbf{0.005} & 1,414                      & 0.011 & 1,379                      & 0.025 & 1,394                       & 0.038  & 1,381                      & 0.330 \\
{\tiny ETTm2}                                                            & 1,380                      & 0.010 & 1,406                      & \textbf{0.007} & 1,398                      & 0.014 & 1,355                      & 0.252 & 1,379                       & 0.031  & 1,321                      & 0.243 \\ \hline
{\tiny \textbf{Average}}                                                 & 1,255                      & 0,054 & 1,273                      & \textbf{0,019} & 1,314                      & 0,035 & 1,242                      & 0,123 & 1,279                       & 0,090  & 1,249                      & 0,182
\end{tabular}
}
\label{table:multi1}
\end{table}

\begin{table}[htbp]
\caption{Univariate RMSE Results}
\centering
{\scriptsize
\setlength\tabcolsep{1.5pt}
\begin{tabular}{c|cc|cc|cc|cc|cc|cc}
\multirow{2}{*}{\textbf{Dataset}} & \multicolumn{2}{c|}{\textbf{Linear}} & \multicolumn{2}{c|}{\textbf{DLinear}} & \multicolumn{2}{c|}{\textbf{NLinear}} & \multicolumn{2}{c|}{\textbf{LSTM}} & \multicolumn{2}{c|}{\textbf{BiLSTM}} & \multicolumn{2}{c}{\textbf{RNN}} \\ \cline{2-13} 
& No VMD      & VMD                 & No VMD       & VMD                 & No VMD           & VMD             & No VMD       & VMD                 & No VMD        & VMD                  & No VMD      & VMD                 \\ \hline
\begin{tabular}[c]{@{}c@{}}{\tiny Banjarbaru PM2.5}\end{tabular}         & 1,011       & 0.620               & 1,12         & 0.703               & 1,188            & 0.908           & 1,216        & \textbf{0.445}      & 1,127         & 0.726                & 1,083       & 0.494               \\
\begin{tabular}[c]{@{}c@{}}{\tiny Batam PM2.5}\end{tabular}              & 1,027       & 0.605               & 1,006        & 0.687               & 1,428            & 0.858           & 1,336        & \textbf{0.398}      & 1,306         & 0.866                & 1,208       & 0.894               \\
\begin{tabular}[c]{@{}c@{}}{\tiny Jakarta PM2.5}\end{tabular}            & 1,042       & \textbf{0.563}      & 1,045        & 0.779               & 1,133            & 0.713           & 1,396        & 1.274               & 1,297         & 0.827                & 1,183       & 0.694               \\
\begin{tabular}[c]{@{}c@{}}{\tiny South Jakarta PM2.5}\end{tabular}      & 1,051       & \textbf{0.662}      & 1,002        & 0.841               & 1,209            & 0.922           & 1,143        & 0.814               & 1,261         & 1.122                & 1,295       & 1.141               \\
\begin{tabular}[c]{@{}c@{}}{\tiny Jambi PM2.5}\end{tabular}              & 0,962       & 0.529               & 0,951        & \textbf{0.521}      & 1,130            & 0.712           & 1,135        & 1.108               & 1,154         & 1.011                & 1,028       & 1.101               \\
\begin{tabular}[c]{@{}c@{}}{\tiny Malang PM2.5}\end{tabular}             & 1,002       & \textbf{0.581}      & 1,007        & 0.855               & 1,166            & 0.953           & 1,221        & 1.222               & 1,172         & 1.260                & 0,999       & 1.161               \\
\begin{tabular}[c]{@{}c@{}}{\tiny Medan PM2.5}\end{tabular}              & 1,000       & 0.545               & 0,994        & 0.714               & 1,079            & 0.962           & 1,164        & 0.572               & 1,080         & \textbf{0.415}       & 1,75        & 0.440               \\
\begin{tabular}[c]{@{}c@{}}{\tiny Samarinda PM2.5}\end{tabular}          & 0,830       & \textbf{0.510}      & 0,832        & 0.652               & 0,988            & 0.634           & 0,897        & 0.844               & 0,935         & 1.435                & 0,853       & 0.796               \\
\begin{tabular}[c]{@{}c@{}}{\tiny Semarang PM2.5}\end{tabular}           & 1,020       & 0.824               & 1,016        & 0.780               & 1,091            & 1.122           & 1,231        & 0.901               & 1,228         & 0.959                & 1,236       & \textbf{0.631}      \\
\begin{tabular}[c]{@{}c@{}}{\tiny M4 Weekly V2}\end{tabular} & 0,591       & 0.631               & 0,592        & 0.753               & 0,636            & 0.613           & 0,682        & 0.307               & 0,635         & \textbf{0.151}       & 0,602       & 0.698               \\
\begin{tabular}[c]{@{}c@{}}{\tiny Central Singapore} \\ {\tiny PM10}\end{tabular}   & 1,056       & \textbf{0.574}      & 1,046        & 0.605               & 1,152            & 0.906           & 1,285        & 0.642               & 1,322         & 0.938                & 1,376       & 0.922               \\
\begin{tabular}[c]{@{}c@{}}{\tiny WindTurbine} \\ {\tiny ActivePower}\end{tabular} & 1,020       & \textbf{0.786}      & 1,025        & 1.022               & 1,398            & 1.295           & 1,421        & 1.940               & 1,424         & 0.865                & 1,392       & 0.919               \\
{\tiny ETTm2 HUFL}                                                          & 1,014       & 0.788               & 1,004        & 0.908               & 1,373            & 0.794           & 1,417        & \textbf{0.744}      & 1,387         & 1.129                & 1,352       & 1.004               \\ \hline
{\tiny \textbf{Average}}                                                    & 0,967       & \textbf{0.619}      & 0,969        & 0.742               & 1,133            & 0.883           & 1,177        & 0.872               & 1,161         & 0.881                & 1,167       & 0.824              
\end{tabular}
}
\label{table:uni1}
\end{table}

\subsection{Forecasting Results}

\textbf{Baseline} we choose other well-known neural network models such as Recurrent Neural Network (RNN), Long Short Term Memory (LSTM), and Bidirectional LSTM (BiLSTM). RNN were developed in 1986 by utilizing Backpropagation Through Time (BPTT) to recognize patterns in sequences of data. RNN were further upgraded to LSTM as an algorithm that can recognize patterns in long sequences of data while maintaining useful information \citep{Sherstinsky_2020}. To enable LSTM to use future contexts of the data, Bidirectional LSTM were developed \citep{Graves2005}. We will also be utilizing VMD together with mentioned neural network models.

\textbf{Results} Multivariate forecasting results were listed in Table \ref{table:multi1}. RMSE values reduction in multivariate predictions were shown after applying VMD with all models having average RMSE values below 0.2. On average, the RMSE decreased by 1.067 to 1.279 when using VMD. The lowest RMSE values were obtained by the DLinear + VMD model in all prediction cases, with an average RMSE of 0.019. The second smallest average RMSE values were held by the NLinear + VMD model with a value of 0.035.

Univariate results were shown in Table \ref{table:uni1}. In most datasets, the use of VMD decomposition methods has proven to significantly enhance the performance of prediction models. However, there are still some cases where the model did not experience a decrease in RMSE, such as in the prediction of WindTurbine ActivePower data with LSTM, which saw an increase in RMSE from 1.421 to 1.940. The best univariate prediction model based on average RMSE across all prediction cases was the Linear model with an average RMSE of 0.619, followed by the DLinear model in second place with an average RMSE of 0.742.

\section{Conclusion And Future Work}

This study demonstrates the effectiveness of Variational Mode Decomposition (VMD) in enhancing the performance of time-series forecasting models by decomposing the original signal into multiple data modes. We integrated VMD with the LTSF-Linear family, which includes Linear, DLinear, and NLinear models, known for their simplicity and competitive edge against Transformer-based models. To rigorously evaluate our approach, we benchmarked it against established neural network models, including RNN, LSTM, and BLSTM, using RMSE as the evaluation metric. The results unequivocally show that VMD significantly reduces RMSE across both univariate and multivariate forecasting tasks. Notably, the Linear + VMD model achieved the best performance on all univariate datasets, while the DLinear + VMD model excelled across all multivariate datasets. These findings highlight the superior performance and robustness of VMD-enhanced linear models in time-series forecasting, providing a promising direction for future research.

Further research should explore optimal forecasting models and try various preprocessing methods. LTSF-Linear has a limited capacity in handling long term time series forecasting but can be further improved by applying various preprocessing methods.

\bibliography{VMDLinear}
\bibliographystyle{VMDLinear}

\end{document}